\pdfoutput=1

\documentclass[11pt]{article}

\usepackage{acl2023}

\usepackage{times}
\usepackage{latexsym}

\usepackage{graphicx}
\usepackage{subfig}
\usepackage{booktabs}
\usepackage{multirow}
\usepackage{svg}
\usepackage{longtable}
\usepackage{tikz}
\usepackage{amsmath}
\usepackage[draft,obeyfinal]{todo}

\usepackage[T1]{fontenc}

\usepackage[utf8]{inputenc}

\usepackage{microtype}
\usepackage[group-separator={,},group-minimum-digits={3}]{siunitx}

\usepackage{inconsolata}
\newsavebox{\tablebox}

\newcommand{\src}{\ensuremath{\mathbf{f}}} 
\newcommand{\trg}{\ensuremath{\mathbf{e}}} 
\newcommand{\newsrc}{\ensuremath{\mathbf{f^{\prime}}}} 
\newcommand{\newtrg}{\ensuremath{\mathbf{e^{\prime}}}} 

\newcommand{\system}[1]{\texttt{{#1}}}

\title{BiSync: A Bilingual Editor for Synchronized Monolingual Texts}

\author{Josep Crego$^{\ddag}$\ $\quad\quad\quad\quad$ Jitao Xu$^{\dag}$\ $\quad\quad\quad\quad$ François Yvon$^{\mathsection}$ \\ \\
$^\ddag$SYSTRAN, 5 rue Feydeau, 75002 Paris, France \\
$^\dag$NetEase YouDao, Beijing, China\\
$^\mathsection$Sorbonne Universit\'e, CNRS, ISIR, F-75005 Paris, France \\
\texttt{firstname.lastname@\{$^{\ddag}$systrangroup.com,$^{\mathsection}$cnrs.fr\} xujt01@rd.netease.com}}

\begin{document}
\maketitle
\begin{abstract}
In our globalized world, a growing number of situations arise where people are required to communicate in one or several foreign languages. In the case of written communication, users with a good command of a foreign language may find assistance from computer-aided translation (CAT) technologies. These technologies often allow users to access external resources, such as dictionaries, terminologies or bilingual concordancers, thereby interrupting and considerably hindering the writing process. In addition, CAT systems assume that the source sentence is fixed and also restrict the possible changes on the target side. In order to make the writing process smoother, we present \system{BiSync}, a bilingual writing assistant that allows users to freely compose text in two languages, while maintaining the two monolingual texts synchronized. We also include additional functionalities, such as the display of alternative prefix translations and paraphrases, which are intended to facilitate the authoring of texts. We detail the model architecture used for synchronization and evaluate the resulting tool, showing that high accuracy can be attained with limited computational resources. The interface and models are publicly available at \url{https://github.com/jmcrego/BiSync} and a demonstration video can be watched \href{https://youtu.be/_l-ugDHfNgU}{\texttt{on YouTube}}.

\end{abstract}

\section{Introduction}

\begin{figure*}[!ht]
\centering
\includegraphics[width=1\textwidth]{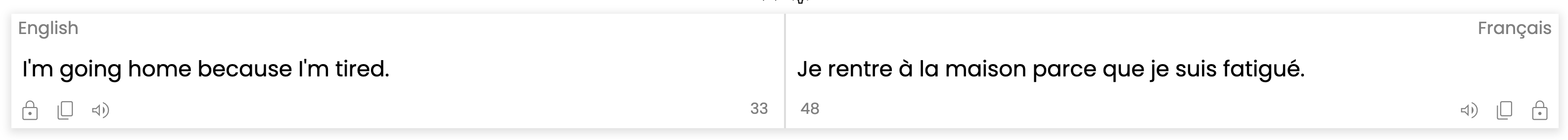}
\caption{User interface of our online bilingual editing system. Users can freely choose the language in which they compose and alternate between text entry boxes. The system automatically keeps the other box in sync.\label{fig:tired}}
\end{figure*}

In today's globalized world, there is an ever-growing demand for multilingual communication. 
To give just a few examples, researchers from different countries often write articles in English, international companies with foreign subsidies need to produce documents in multiple languages, research institutions communicate in both English and the local language, etc. However, for many people, writing in a foreign language (L2) other than their native language (L1) is not an easy task.

With the significant advances in machine translation (MT) in the recent years, in particular due to the tangible progress in neural machine translation (NMT, \citealp{Bahdanau15neural,Vaswani17attention}), MT systems are delivering usable translations in an increasing number of situations. However, it is not yet realistic to rely on NMT technologies to produce high quality documents, as current state-of-the-art systems have not reached the level where they could produce error-free translations. Also, fully automatic translation does not enable users to precisely control the output translations (e.g.\ with respect to style, formality, or term use). Therefore, users with a good command of L2, but not at a professional level, can find help from existing computer-assisted language learning tools or computer-assisted translation (CAT) systems. These tools typically provide access to external resources such as dictionaries, terminologies, or bilingual concordancers \citep{Bourdaillet11transsearch} to help with writing. However, consulting external resources causes an interruption in the writing process due to the initiation of another cognitive activity, even when writing in L1 \citep{Leijten14writing}. Furthermore, L2 users tend to rely on L1 \citep{Wolfersberger03l1} to prevent a breakdown in the writing process \citep{Cumming89writing}. To this end, several studies have focused on developing MT systems that ease the writing of texts in L2 \citep{Koehn10enabling,Huang12transahead,Venkatapathy12smtdriven,Chen12flow,Liu16agreement}.

However, existing studies often assume that users can decide whether the provided L2 texts precisely convey what they want to express. Yet, for users who are not at a professional level, the evaluation of L2 texts may not be so easy. To mitigate this issue, researchers have also explored round-trip translation (RTT), which translates the MT output in L2 back to L1 in order to evaluate the quality of L2 translation \citep{Moon20revisiting}. Such studies suggest that it is then helpful to augment L2 writing with the display of the corresponding synchronized version of the L1 text, in order to help users verify their composition. In such settings, users can obtain synchronized texts in two languages, while only making an effort to only compose in one. 

A bilingual writing assistant system should allow users to write freely in both languages and always provide synchronized monolingual texts in the two languages. However, existing systems do not support both functionalities simultaneously. The system proposed by \citet{Chen12flow} enables free composition in two languages, but only displays the final texts in L2. Commercial MT systems like Google,\footnote{\url{https://translate.google.com/}} DeepL\footnote{\url{https://www.deepl.com/translator}} and SYSTRAN\footnote{\url{https://www.systran.net/en/translate/}} always display texts in both languages, but users can only modify the source side, while the target side is predicted by the system and is either locked or can only be modified with alternative translations proposed by the system. CAT tools, on the contrary, assume the source sentence is fixed and only allow edits on the target side.
 
In this paper, we present \system{BiSync}, a bilingual writing assistant aiming to extend commercial MT systems by letting users freely alternate between two languages, changing the input text box at their will, with the goal of authoring two equally good and semantically equivalent versions of the text.

\begin{figure*}[!ht]
\centering
\subfloat{\includegraphics[width=1\textwidth]{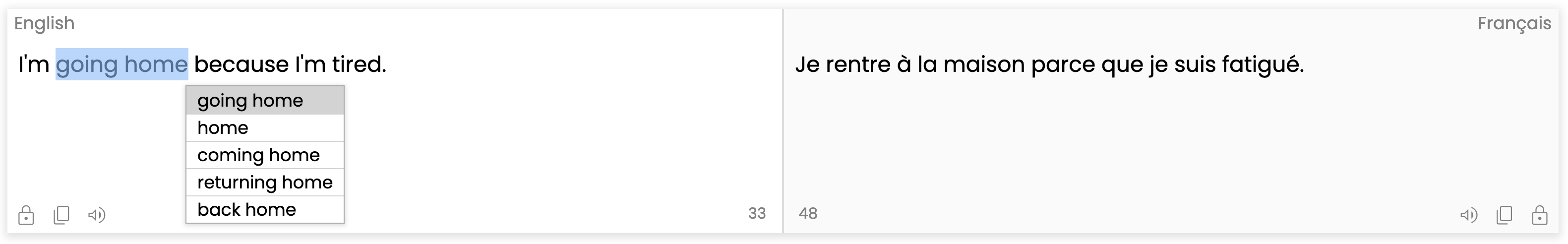}} \\
\subfloat{\includegraphics[width=1\textwidth]{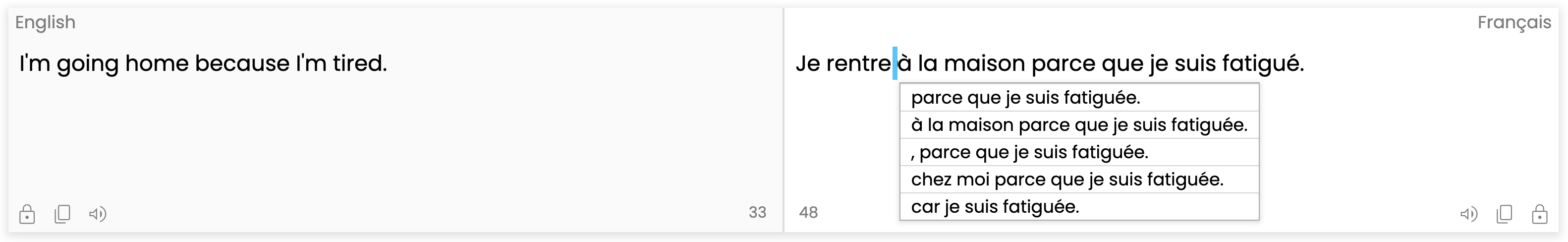}}
\caption{\system{BiSync} editor displaying paraphrases (top) and translation alternatives for a given prefix (bottom).} \label{fig:menu}
\end{figure*}

\section{\system{BiSync} Text Editor}
\label{sec:bisync}

In this work, we are interested in a writing scenario that broadens the existing commercial online translation systems. We assume that the user wants to edit or revise a text simultaneously in two languages. See Figure~\ref{fig:tired} for a snapshot of our \system{BiSync} assistant. Once the text is initially edited in one language, the other language is automatically synchronized so that the two entry boxes always contain mutual translations. In an iterative process, and until the user is fully satisfied with the content, texts are revised in either language, triggering automatic synchronizations to ensure that both texts remain mutual translations. The next paragraphs detail the most important features of our online \system{BiSync} text editor. 

\paragraph{Bidirectional Translations} 

The editor allows users to edit both text boxes at their will. This means that the underlying synchronization model has to perform translations in both directions, as the role of the source and target texts are not fixed and can change over time.

\paragraph{Synchronization}

This is the most important feature of the editor. It ensures that the two texts are always translations of each other. As soon as one text box is modified, \system{BiSync} synchronizes the other box. To enhance the user experience, the system waits a few seconds (delay) before the synchronization takes place. When a text box is modified, the system prevents the second box from being edited until the synchronization has been completed. Users can also disable the synchronization process, using the "freeze" button (\raisebox{-5pt}{\tikz{\node at (0,0){\includegraphics[width=1em]{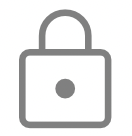}};}}). In this case, the frozen text will not be synchronized (modified). Changes are only allowed in the unfrozen text box. This is the standard \textsl{modus operandi} of most commercial translation systems that consider the input text as frozen, allowing only a limited number of edits in the translation box.

\paragraph{Prefix Alternatives}

The editor can also provide several translation alternatives for a given sentence prefix. When users click just before a word $w$ in a synchronized sentence pair, the system displays the most likely alternatives that can complete the translation starting from the word $w$ in a drop-down menu. Figure~\ref{fig:menu} (bottom) illustrates this functionality, where translation alternatives are displayed after the prefix "\textit{Je rentre}", in the context of the English sentence "\textit{I'm going home because I'm tired}". In the example in Figure~\ref{fig:menu} (bottom), the user clicked right before the French word "\textit{à}".

\paragraph{Paraphrase Alternatives} 

Another important feature of our \system{BiSync} editor is the ability to propose edits for sequences of words at arbitrary positions in both text boxes. Figure~\ref{fig:menu} (top) illustrates this scenario, where paraphrases for the English segment "\textit{going home}" are displayed in the context "\textit{I'm ... because I'm tired}" and given the French sentence "\textit{Je rentre à la maison parce que je suis fatigué}". Such alternatives are triggered through the selection of word sequences in either text box. 

\paragraph{Other Features}

Like most online translation systems, \system{BiSync} has a "listen" button that uses a text-to-speech synthesizer to read the content in the text box, a "copy" button that copies the content to the clipboard. It also displays the number of characters written in each text box. Figures~\ref{fig:tired} and \ref{fig:menu} illustrate these features. 

\paragraph{Settings}

\begin{figure}[!ht]
\center
\includegraphics[width=0.73\columnwidth]{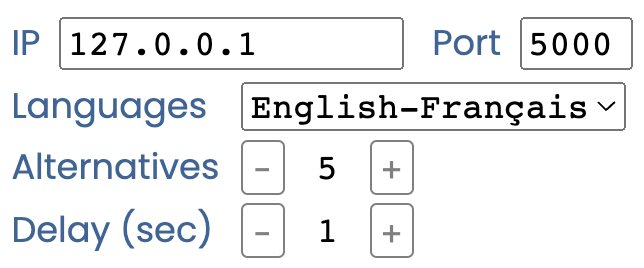}
\caption{Default \system{BiSync} settings.\label{fig:settings}}
\end{figure}

The "gear" button (\raisebox{-5pt}{\tikz{\node at (0,0){\includegraphics[width=1em]{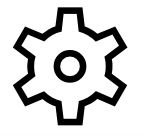}};}}) pops up several system parameters that can be configured by users: The "IP" and "Port" fields identify the address where the \system{BiSync} model is launched and waits for translation requests. The "Languages" field indicates the pair of languages that the model understands and is able to translate. "Alternatives" denotes the number of hypotheses that the model generates and that are displayed by the system. "Delay" sets the number of seconds the system waits after an edit takes place before starting the synchronization. The countdown is reset each time a revision is made. Figure~\ref{fig:settings} displays \system{BiSync} settings with default parameters.

\section{Under the Hood: the \system{BiSync} Model}

Given the features that we would like to offer in the \system{BiSync} text editor, we are interested in an end-to-end model producing: (a) translations in both directions; (b) translations from scratch, with only one text box filled in; (c) updates of existing translations, so that small changes in one text box result in small updates in the other text box; (d) paraphrases and alternatives in context.

\begin{figure*}[!ht]
\center
\begin{tabular}{l|ll|ll}
\hline
$Update$ & \textcolor{gray}{\src} & \newsrc & \trg & \newtrg \\
\specialrule{2pt}{1pt}{1pt} 
$Ins$ & \textcolor{gray}{The cat}                   & The \textbf{white} cat & Le chat           & Le chat blanc \\ 
$Del$ & \textcolor{gray}{The cat \textbf{is} white} & The white cat          & Le chat est blanc & Le chat blanc \\ 
$Sub$ & \textcolor{gray}{The \textbf{black} cat}    & The \textbf{white} cat & Le chat noir      & Le chat blanc \\ 
\hline
\end{tabular}
\caption{Source sentences \src{} when updated (\newsrc) by means of insertion ($Ins$), deletion ($Del$) and substitution ($Sub$) and their corresponding translations (\trg{} and \newtrg). Source sentences \src{} are not employed by the models of this work.\label{fig:ids}}
\end{figure*}

We consider a pair of parallel sentences $(\src, \trg)$ and a sentence \newsrc{} as an update of \src. The objective is to generate the sentence \newtrg{} that is parallel to \newsrc{} while remaining as close as possible to \trg. Three types of update are distinguished. Figure~\ref{fig:ids} displays an example for each update type:
\begin{itemize}
    \item \textbf{Insertion}: adding one or more consecutive words at any position in \src; 
    \item \textbf{Deletion}: removing one or more consecutive words at any position in \src; 
    \item \textbf{Substitution}: replacing one or more consecutive words at any position in \src{} by one or more consecutive words. 
\end{itemize}
Note that in practice, training such models requires triplets $(\newsrc, \trg, \newtrg)$, as sentences \src{} are not used by the models studied in this work.

Inspired by \citet{Xiao22bitiimt}, we integrate several control tokens into the source-side of training examples of a standard NMT model to obtain the desired results. This approach is straightforward and does not require to modify NMT architectures or decoding algorithms. Therefore, our integration of control tokens is model-agnostic and can be applied to any NMT architecture. Several tokens are used to indicate the target language (\verb|<en>|, \verb|<fr>|) and the update type  (\verb|<ins>|, \verb|<del>|, \verb|<sub>|). A special token \verb|<gap>| indicates a sequence of masked tokens used to generate paraphrases. 

Training examples are build following the next three patterns:

\begin{center}
\begin{tabular}{|l|l|}
\hline
    \newsrc{} \verb|<lang>| & \newtrg \\
\hline
\end{tabular}
\end{center}

The first pattern refers to a regular translation task (\system{TRN}), and is used when translating from scratch, without any initial sentence pair $(\src, \trg)$, as in the following example:
\begin{center}
\begin{tabular}{l|l}
\textit{The white cat } \verb|<fr>| & \textit{Le chat blanc} \\
\end{tabular}
\end{center}
where only the target language tag is appended to the end of the source sentence.

\begin{center}
\begin{tabular}{|l|l|}
\hline
    \newsrc{} \verb|<lang>| \trg{} \verb|<update>| & \newtrg \\
\hline
\end{tabular}
\end{center}

The second pattern corresponds to an update task (\system{INS}, \system{DEL} or \system{SUB}) to be used for resynchronizing an initial sentence pair $(\src, \trg)$ after changing $\src$ into $\newsrc$, as shown in the following examples:
\begin{flushleft}
\begin{lrbox}{\tablebox}
  \begin{tabular}{l|l}
   \textit{The white cat} \verb|<fr>| \textit{Le chat} \verb|<ins>| & \textit{Le chat blanc} \\
   \textit{The white cat} \verb|<fr>| \textit{Le chat est blanc} \verb|<del>| & \textit{Le chat blanc} \\
  \textit{The white cat} \verb|<fr>| \textit{Le chat noir} \verb|<sub>| & \textit{Le chat blanc} \\
  \end{tabular}
\end{lrbox}
\resizebox{\columnwidth}{!}{\usebox{\tablebox}}
\end{flushleft}
\vspace{0.2cm}
\noindent where the edited source sentence $\newsrc =\ $[\textit{The white cat}] is followed by the target language tag \verb|<fr>|, the initial target sentence \trg{}, and a tag indicating the edit type that updates the initial source sentence \src. 

\begin{center}
\begin{tabular}{|l|l|}
\hline
    \src{} \verb|<lang>| $\trg_g$ & $\trg_G$ \\
\hline
\end{tabular}
\end{center}

The third pattern corresponds to a bilingual text infilling task (\system{BTI}, \citealp{Xiao22bitiimt}). The model is trained to predict the tokens masked in a target sentence $\trg_g$ in the context of the source sentence $\src$:

\begin{center}
\begin{tabular}{l|l}
\textit{The white cat} \verb|<fr>| \textit{Le} \verb|<gap>| \textit{blanc} & \textit{chat} \\
\end{tabular}
\end{center}

\noindent where $\trg_g =\ $ [\textit{Le} \verb|<gap>| \textit{blanc}] is the target sentence with missing tokens to be predicted. The model only generates the masked tokens $\trg_G =\ $[\textit{chat}]. 

\subsection{Synthetic Data Generation\label{ssec:sdg}}

While large amounts of parallel bilingual data $(\newsrc, \newtrg)$ exist for many language pairs, the triplets required to train our model are hardly available. We therefore study ways to generate synthetic triplets of example $(\newsrc, \trg, \newtrg)$ from parallel data $(\newsrc, \newtrg)$ for each type of task (\system{INS}, \system{DEL}, \system{SUB} and \system{BTI}) introduced above. 

\paragraph{Insertion}
We build examples of initial translations \trg{} for \system{INS} by randomly dropping a segment from the updated target \newtrg. 
The length of the removed segment is also randomly sampled with a maximum length of $5$ tokens. 
We also impose that the overall ratio of removed segment does not exceed $0.5$ of the length of \newtrg.

\paragraph{Deletion}
Simulating deletions requires the initial translation \trg{} to be an extension of the updated target \newtrg. 
To obtain extensions, we employ a NMT model enhanced to fill in gaps (\system{fill-in-gaps}). 
This model is a regular encoder-decoder Transformer model trained with a balanced number of regular parallel examples (\system{TRN}) and paraphrase examples (\system{BTI}) as detailed in the previous section. 
We extend training examples $(\newsrc, \newtrg)$ with a \verb|<gap>| token inserted in a random position in \newtrg{} and use \system{fill-in-gaps} to decode these training sentences, as proposed in \citep{Xu22bilingual}. In response, the model predicts tokens that best fill the gap. For instance:

\begin{center}
\begin{tabular}{l}
\textit{The white cat} \verb|<fr>| \textit{Le chat} \verb|<gap>| \textit{blanc} $\leadsto$ \textit{est}
\end{tabular}
\end{center}

\noindent the target extension is therefore $\trg = $ [\textit{Le chat est blanc}].

\paragraph{Substitution}
Similar to deletion, substitution examples are obtained using the same \system{fill-in-gaps} model. A random segment is masked from \newtrg, which is then filled by the model. In inference, the model computes an $n$-best list of substitutions for the mask, and we select the most likely sequence that is not identical to the masked segment. For instance:

\vspace{0.2cm}
\textit{The white cat} \verb|<fr>| \textit{Le chat} \verb|<gap>| $\leadsto$ [\textit{blanc; bleu; clair; blanche; ...}] 
\vspace{0.2cm}

\noindent the target substitution is $\trg = $ [\textit{Le chat bleu}].

Note that extensions and substitutions generated by \system{fill-in-gaps} may be ungrammatical. For instance, the proposed substitution $\trg = $ [\textit{Le chat blanche}] has a gender agreement error. The correct adjective should be "\textit{blanc}" (masculine) instead of "\textit{blanche}" (feminine). This way, the model always learns to produce grammatical \newtrg{} sentences parallel to \newsrc.

\paragraph{Paraphrase}
Given sentence pairs $(\src, \trg)$, we generate $\trg_g$ by masking a random segment from the initial target sentence $\trg$. 
The length of the masked segment is also randomly sampled with a maximum length of $5$ tokens. 
The target side of these examples ($\trg_G$) only contains the masked token(s).

\section{Experiments}

\subsection{Datasets}

To train our English-French (En-Fr) models we use the official WMT14 En-Fr corpora\footnote{\url{https://www.statmt.org/wmt14}} as well as the OpenSubtitles corpus\footnote{\url{https://opus.nlpl.eu/OpenSubtitles-v2018.php}} \citep{Lison16opensubtitles}. A very light preprocessing step is performed to normalize punctuation and to discard examples exceeding a length ratio $1.5$ and a limit of $[1,250]$ measured in words. Statistics of each corpus is reported in Table~\ref{tab:traindata}.

\begin{table}[!ht]
\center
\begin{tabular}{lr}
\hline
$Corpora$ & $\# Sentences$ \\
\specialrule{2pt}{1pt}{1pt}
Europarl v7                & \num{2007723} \\
Commoncrawl & \num{3244152} \\
UN                   & \num{12886831} \\ 
News Commentary            & \num{183251} \\
Giga French-English  & \num{22520376} \\
Open Subtitles v18         & \num{57123540} \\
\hline
Total                      & \num{97965873} \\
\hline
\end{tabular}
\caption{Statistics of training corpora.\label{tab:traindata}}
\end{table}

For testing, we used the official newstest2014 En-Fr test set made available for the same WMT14 shared task containing \num{3003} sentences. All our data is tokenized using OpenNMT tokenizer.\footnote{\url{https://github.com/OpenNMT/Tokenizer}} We learn a joint Byte Pair Encoding \citep{Sennrich16BPE} over English and French training data with $32$k merge operations.

The training corpora used for learning our model consist of well-formed sentences. Most sentences start with a capital letter and end with a punctuation mark. However, our \system{BiSync} editor expects also incomplete sentences, when synchronization occurs before completing the text. To train our model to handle this type of sentences, we lowercase the first character of sentences and remove ending punctuation in both source and target examples with a probability set to $0.05$.

\subsection{Experimental Settings}

Our \system{BiSync} model is built using the Transformer architecture \citep{Vaswani17attention} implemented in \texttt{OpenNMT-tf}\footnote{\url{https://github.com/OpenNMT/OpenNMT-tf}} \citep{Klein17opennmt}. More precisely, we use the following set-up: embedding size: \num{1024}; number of layers: $6$; number of heads: $16$; feedforward layer size: \num{4096}; and dropout rate: $0.1$. We share parameters for input and output embedding layers \citep{Press17using}. We train our models using Noam schedule \citep{Vaswani17attention} with \num{4000} warm-up iterations. Training is performed over a single V100 GPU during $500$k steps with a batch size of \num{16384} tokens per step. We apply label smoothing to the cross-entropy loss with a rate of $0.1$. Resulting models are built after averaging the last ten saved checkpoints of the training process. For inference, we use \texttt{CTranslate2}.\footnote{\url{https://github.com/OpenNMT/CTranslate2}} It implements custom run-time with many performance optimization techniques to accelerate decoding execution and reduce memory usage of models on CPU and GPU. We also evaluate our model with weight quantization using 8-bit integer (\system{int8}) precision, thus reducing model size and accelerating execution compared to the default 32-bit float (\system{float}) precision.

\section{Evaluation\label{sec:evaluation}}

We evaluate the performance of our synchronization model \system{BiSync} compared to a baseline translation model with exactly the same characteristics but trained only on the \system{TRN} task over bidirectional parallel data (\system{base}). We report performance with BLEU score \citep{Papineni02bleu} implemented in SacreBLEU\footnote{\url{https://github.com/mjpost/sacrebleu}. Signature: nrefs:1|case:mixed|eff:no|tok:13a|smooth:exp| version:2.0.0} \citep{Post18sacrebleu} over concatenated En-Fr and Fr-En test sets. For tasks requiring an initial target $\trg$, we synthesize $\trg$ from $(\newsrc,\newtrg)$ pairs following the same procedures used for generating the training set (see details in Section~\ref{sec:bisync}).

Table~\ref{tab:eval_onmt} reports BLEU scores for our two systems on all tasks. The \system{base} system is only trained to perform regular translations (\system{TRN}) for which it obtains a BLEU score of $36.0$, outperforming \system{BiSync}, which is trained to perform all tasks. This difference can be explained by the fact that \system{BiSync} must learn a larger number of tasks than \system{base}. When performing \system{INS}, \system{DEL} and \system{SUB} tasks, \system{BiSync} vastly outperforms the results of \system{TRN} task as it makes good use of the initial translation $\trg$. When we use \system{BiSync} to generate paraphrases of an initial input (\system{BTI}), we obtain a higher BLEU score of $42.6$ than the regular translation task (\system{TRN}, $34.9$). This demonstrates the positive impact of using target side context for paraphrasing.

\begin{table}[htbp]
\center
\begin{tabular}{l|ccccc}
\hline
BLEU & \system{TRN} & \system{INS} & \system{DEL} & \system{SUB} & \system{BTI}\\
\specialrule{2pt}{1pt}{1pt}
\system{base}   & $36.0$ & -      & -      & -      & -      \\
\system{BiSync} & $34.9$ & $87.9$ & $95.5$ & $78.2$ & $42.6$ \\
\hline
\end{tabular}
\caption{BLEU scores over concatenated En-Fr and Fr-En test sets for all tasks. \label{tab:eval_onmt}}
\end{table}

Next, we evaluate the ability of our \system{BiSync} model to remain close to an initial translation when performing synchronization. Note that for a pleasant editing experience, synchronization should introduce only a minimum number of changes. Otherwise, despite re-establishing synchronization, additional changes may result in losing updates previously performed by the user. To evaluate this capability of our model, we take an initial translation $(\src, \trg)$ and introduce a synthetic update (say $\newsrc$) as detailed in Section~\ref{ssec:sdg}. This update leads to a new synchronization that transforms $\trg$ into $\newtrg$. We would like $\newtrg$ to remain as close as possible to $\trg$. Table~\ref{tab:distance} reports TER scores \citep{Snover06study} between $\trg$ and $\newtrg$ computed by SacreBLEU.\footnote{Signature: nrefs:1|case:lc|tok:tercom|norm:no|punct:yes |asian:no|version:2.0.0} These results indicate that \system{BiSync} produces synchronizations significantly closer to initial translations than those produced by \system{base}. This also confirms the findings of \citet{Xu22bilingual}.

\begin{table}[htbp]
\center
\begin{tabular}{l|rrr}
\hline
TER $\downarrow$ & \system{INS} & \system{DEL} & \system{SUB} \\
\specialrule{2pt}{1pt}{1pt}
\system{base}   & $36.5$ & $43.4$ & $34.9$ \\
\system{BiSync} & $3.3$  & $5.5$  & $5.0$  \\
\hline
\end{tabular}
\caption{TER scores between \trg{} and \newtrg{} issued from different update types. En-Fr and Fr-En test sets are concatenated. \label{tab:distance}}
\end{table}

Finally, Table~\ref{tab:eval_ct2} reports inference efficiency for our \system{BiSync} model using \texttt{CTranslate2}. We indicate differences in model (\textit{Size} and \textit{Speed}) for different quantization, device, batch size and number of threads. Lower memory requirement and higher inference speed can be obtained by using quantization set to int-8 for both GPU and CPU devices, in contrast to float-32. When running on CPUs, additional speedup is obtained with multithreading. Comparable BLEU scores are obtained in all configurations. Note that for the tool presented in this paper, we must retain single batch size and single thread results (bold figures), since synchronization requests are produced for isolated sentences. Therefore, they cannot take advantage of using multiple threads and large batches.
\begin{table}[htbp]
\center
\resizebox{\columnwidth}{!}{ 
\begin{tabular}{clcc|rr}
\hline
\textit{Quant} & \textit{Dev} & \textit{BS} & \textit{Threads} & \textit{Size} & \textit{Speed} \\
\specialrule{2pt}{1pt}{1pt}
\multirow{4}{*}{\system{float}} & GPU & $64$ & $1$ & \multirow{4}{*}{$232$M} & \num{10267} \\
                                & GPU & $1$  & $1$ &  & $\mathbf{650}$ \\
                                & CPU$^1$ & $64$ & $8\times4$ &  & \num{2007}\\
                                & CPU$^2$ & $1$  & $1$   &  & $\mathbf{48}$ \\
\hline
\multirow{4}{*}{\system{int8}}  & GPU & $64$ & $1$   & \multirow{4}{*}{$59$M} & \num{12918} \\
                                & GPU & $1$  & $1$ &  & $\mathbf{738}$\\
                                & CPU$^1$ & $64$ & $8\times4$ &  & \num{2666} \\
                                & CPU$^2$ & $1$  & $1$   &  & $\mathbf{118}$ \\
\hline
\end{tabular}
}
\caption{Inference \textit{Speed} and model \textit{Size} when decoding test sets with several settings: quantization (\textit{Quant}), device (\textit{Dev}), batch size (\textit{BS}) and number of \textit{Threads}. Decoding beam size is set to $3$. \textit{Speed} is measured in tokens/second. GPU is a single V100 GPU with 32Gb memory. CPU$^1$ has 32 cores with 86Gb memory and CPU$^2$ is an Intel i7-10850H with 32Gb memory. \label{tab:eval_ct2}}
\end{table}

\section{Conclusion and Further Work}

In this paper, we presented \system{BiSync}, a bilingual writing assistant system that allows users to freely compose text in two languages while always displaying the two monolingual texts synchronized with the goal of authoring two equally good versions of the text. Whenever users make revisions on either text box, \system{BiSync} takes into account the initial translation and reduces the number of changes needed in the other text box as much as possible to restore parallelism. \system{BiSync} also assists in the writing process by suggesting alternative reformulations for word sequences or alternative translations based on given prefixes. The synchronization process applies several performance optimization techniques to accelerate inference and reduce the memory usage with no accuracy loss, making \system{BiSync} usable even on machines with limited computing power.

In the future, we plan to equip \system{BiSync} with a grammatical error prediction model and a better handling of prior revisions: the aim is to enable finer-grained distinction between parts that the system should modify and parts that have already been fixed or that should remain unchanged. Last, we would like to perform user studies to assess the division of labor between users and \system{BiSync} in an actual bilingual writing scenario.

\section*{Acknowledgements}

We would like to thank the anonymous reviewers for their valuable suggestions and comments. This work was performed while the last two authors where affiliated with LISN, CNRS (Orsay). Jitao Xu has been partly funded by SYSTRAN and by a grant (Transwrite) from the Région Ile-de-France. He was granted access to the HPC resources of IDRIS under the allocation 2022-[AD011011580R2] made by GENCI.

\bibliography{biblio}
\bibliographystyle{acl_natbib}

\end{document}